\documentclass[10pt,twocolumn,letterpaper]{article}

\usepackage{cvpr}
\usepackage{times}
\usepackage{epsfig}
\usepackage{graphicx}
\usepackage{amsmath}
\usepackage{amssymb}

\usepackage{amsmath}
\usepackage{subfigure}
\usepackage{epsfig}
\usepackage{graphicx}
\usepackage{amsmath}
\usepackage{amssymb}
\usepackage[ruled]{algorithm2e}
\usepackage{listings}
\usepackage{textcomp,booktabs}
\usepackage{color}
\usepackage{xcolor}
\usepackage{colortbl}
\usepackage{tabu}
\usepackage{threeparttable}

\usepackage{bbding}
\usepackage{pifont}
\usepackage{wasysym}
\usepackage{amssymb}
\usepackage{url}

\newcommand*{\affaddr}[1]{#1} 
\newcommand*{\affmark}[1][*]{\textsuperscript{#1}}
\newcommand*{\email}[1]{\texttt{#1}}
\renewcommand*{\email}{\small}

\usepackage[pagebackref=true,breaklinks=true,letterpaper=true,colorlinks,bookmarks=false]{hyperref}

\cvprfinalcopy 

\def\cvprPaperID{4577} 

\ifcvprfinal\fi

\begin{document}

\title{Fast Template Matching and Update for Video Object Tracking and Segmentation}

\author{%
	 Mingjie Sun\affmark[1,2],\quad Jimin Xiao\affmark[1,\thanks{corresponding author}],\quad Eng Gee Lim\affmark[1],\quad Bingfeng Zhang\affmark[1,2],\quad Yao Zhao\affmark[3]\\
	\affaddr{\affmark[1]XJTLU},\quad
	\affaddr{\affmark[2]University of Liverpool},\quad
	\affaddr{\affmark[3]Beijing Jiaotong University}\\
	\email{mingjie.sun@liverpool.ac.uk},\quad
	\email{\{jimin.xiao, enggee.lim, bingfeng.zhang\}@xjtlu.edu.cn},\quad
	\email{yzhao@bjtu.edu.cn}
}

\maketitle

\begin{abstract}
\footnotetext[1]{The work was supported by National Natural Science Foundation of China under 61972323,61532005 and U1936212, and Key Program Special Fund in XJTLU under KSF-T-02, KSF-P-02.
		
\url{https://github.com/insomnia94/FTMU}}
In this paper, the main task we aim to tackle is the multi-instance semi-supervised video object segmentation across a sequence of frames where only the first-frame box-level ground-truth is provided. Detection-based algorithms are widely adopted to handle this task, and the challenges lie in the selection of the matching method to predict the result as well as to decide whether to update the target template using the newly predicted result. The existing methods, however, make these selections in a rough and inflexible way, compromising their performance. To overcome this limitation, we propose a novel approach which utilizes reinforcement learning to make these two decisions at the same time. Specifically, the reinforcement learning agent learns to decide whether to update the target template according to the quality of the predicted result. The choice of the matching method will be determined at the same time, based on the action history of the reinforcement learning agent. Experiments show that our method is almost 10 times faster than the previous state-of-the-art method with even higher accuracy (region similarity of 69.1\% on DAVIS 2017 dataset).
\end{abstract}

\section{Introduction}

Multi-instance semi-supervised video object segmentation (VOS) is an important computer vision task, serving as the basis of many other related tasks including scene understand, video surveillance and video editing. The task of VOS is to produce instance segmentation masks for each frame in a video sequence where the first-frame ground-truth is provided in advance. It turns out to be a challenging task especially in the situations of deformation, motion blur, illumination change, background clutter, and so on.

\begin{figure}[t]
	\begin{center}
		\includegraphics[width=0.9\linewidth]{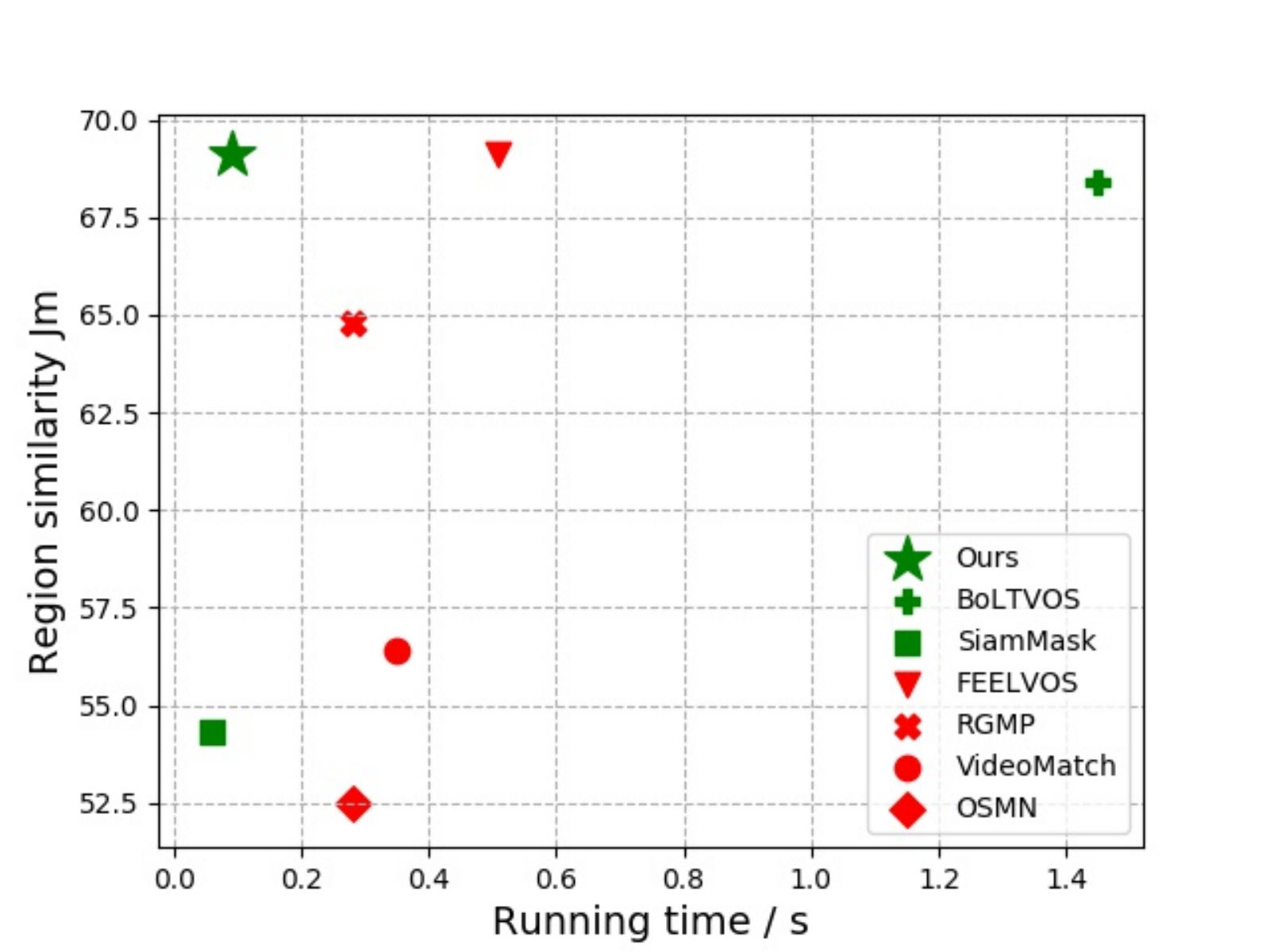}
	\end{center}
	\caption{Speed-accuracy trade-off for various multi-instance semi-supervised VOS methods on DAVIS 2017 dataset. Methods in green only rely on the first-frame box-level ground-truth while methods in red rely on the first-frame pixel-level ground-truth.}
	\label{tradeoff}
\end{figure}

There are two common ways to provide the first-frame ground-truth, including masks and bounding boxes. Providing the first-frame masks is a conventional way, which has been widely adopted nowadays \cite{wug2018fast,voigtlaender2019feelvos}. Although these methods have already achieved good performance with the pixel-level accurate target object information, it turns to a hard task to utilize these methods to solve practical VOS problems, especially when numerous video sequences need to be processed in a short time, because annotating pixel-level ground-truth masks for each video sequence is time-consuming. To overcome this problem, inspired by the rapid progress in the task of video object tracking (VOT) at bounding box level, some works attempt to rely on the first-frame bounding boxes to provide target object information instead of using the first-frame masks, which dramatically accelerates the annotation process and increases scalability.

This kind of acceleration is, however, built on the sacrifice of ground-truth ``accuracy''. The reason is that some background area will be incorporated into the bounding box as well, which greatly increases the difficulty of the VOS task. In this way, in order to adapt to the characteristics of box-level ground-truth, most existing methods relying on the first-frame bounding boxes adopt the detection-based algorithms. Generally, these detection-based methods include three steps. The first step is to conduct object detection on the whole frame to generate the proposals of all possible objects using the region proposal network \cite{ren2015faster}. The second step is to do the matching process between the target object and all candidate proposals to find the ``correct'' proposal. The third step is to do the salience segmentation on the 	``correct'' proposal to generate the final segmentation result. However, existing methods relying on the first-frame bounding boxes \cite{wang2019fast,voigtlaender2019boltvos} are less performing than the methods relying on first-frame masks both in terms of running speed and accuracy. 

First, in terms of the running speed, we observe that most existing detection-based algorithms spend too much time on the matching process (e.g., 1.425s for matching process and 0.025s for segmentation process in \cite{voigtlaender2019boltvos}), using several time-consuming networks to evaluate the appearance similarity, like the re-identification network \cite{sun2019ovsnet} and the siamese style network \cite{voigtlaender2019boltvos}. We observe that, for most video sequences, fast matching according to the intersection over union (IOU) between candidate proposals of the current frame and the obtained previous frame's bounding box or segmentation mask can also lead to acceptable performance, because the target object normally moves or changes slowly in two successive frames. However, simple IOU-based matching methods sometimes lose the target, especially when the target disappears out of the sight, and then reappears at a different location. Therefore, neither simple appearance-based matching nor simple IOU-based matching is the best solution for this task.

Second, the major constriction of VOS accuracy is its rough way to update the target template. The target template, containing the target's latest information including appearance, location and so on, plays an essential role in the matching process. Among all candidate proposals of the current frame, the proposal with the highest similarity with the target template will be selected. In this way, whether the correct proposal can be selected is determined by the quality of the target template as well as its update mechanism. Existing methods, however, simply replace the target template with the newly predicted result after one frame is finished, regardless of the correctness of the obtained result. Therefore, error will be gradually introduced into the target template, causing a great accuracy decline. 

To achieve a better balance between accuracy and speed, a ``smart switch'' is required to make two significant decisions, including adopting which matching method, IOU-based matching or appearance-based matching, and whether to update the target template or not. To tackle this problem, we formalize it as a conditional decision-making process where only one simple reinforcement learning (RL) agent is employed to make decisions in a flexible way. As can be observed from Figure \ref{tradeoff}, provided with the optimal matching method and updating mechanism, our algorithm can be dramatically accelerated without losing targets even in some difficult frames, leading to a higher accuracy against previous state-of-the-art methods. Specifically, the running speed of our method is approximately 10 times faster than the previous state-of-the-art method.

To sum up, most video object tracking and segmentation algorithms consist of three steps. The first step is to conduct the instance segmentation on the current frame to generate a pool of candidate proposals. The second step is to conduct the matching process to find the correct one as the final result among all candidate proposals according to the target template information. The third step is to entirely replace the target template using the prediction of the current frame. In this paper, 
as we find the first step does not greatly affect the final result, our novelty lies in the improvement of the second and third steps:
\begin{itemize}
	\item 
	To improve the second step, our method provides a simple way to trade off between running speed and accuracy by selecting the matching method (IOU-based matching or appearance-based matching). The choice of the matching method is determined by the action history of the RL agent, which dramatically reduces the running time of our method. 
	
	\item 
	To improve the third step, as we observe the importance of the target template update mechanism to avoid drift, we argue that some predicted results with terrible quality should be discarded, and the target template should be kept unchanged in this situation. Specifically, we adopt a RL agent to make the decision on whether to update the target template or not according to the quality of the predicted result, which effectively prevents the drift problem and boots the accuracy of our method.
	
	\item 
	The proposed approach has been validated on both VOS and VOT datasets, including DAVIS 2017, DAVIS 2016, SegTrack V2, Youtube-Object and VOT 2018 long-term datasets. Our method is approximately 10 times faster than the previous state-of-the-art method and achieve a higher mean region similarity at the same time. The new state-of-the-art mean region similarity is obtained on several datasets including DAVIS 2017 (69.1\%), SegTrack V2(79.2\%) and Youtube-Object (79.3\%).
\end{itemize}

\section{Related Work}

\subsection{Video Object Segmentation}
VOS can be classified into three different categories including unsupervised VOS \cite{tokmakov2017learning}, interactive VOS \cite{chen2018blazingly,wang2005interactive} and semi-supervised VOS \cite{voigtlaender2019feelvos,chen2018blazingly}.

Unsupervised VOS is the task where no first-frame annotation is available at all. In \cite{song2018pyramid}, concatenated pyramid dilated convolution feature map is utilized to improve the final accuracy. Interactive VOS allows user annotation. In \cite{chen2018blazingly}, an embedding model is trained to tell if two pixels belong to the same object, which proves to be efficient for this task. 

Currently, semi-supervised VOS, where the first-frame ground-truth is provided, is still the main battlefield of the VOS tasks. The most common approach is to use the first-frame ground-truth to fine-tune the general segmentation network \cite{caelles2017one}. To adjust to the object appearance variation, in \cite{voigtlaender2017online}, the segmentation network will be updated during the test time. To overcome the speed shortcoming of the online updating, in \cite{xiao2019online}, a meta learning model is utilized to speed up the process of online updating virtually without a reduction on the accuracy. To overcome the lack of the training data, it is proposed to utilize static images to generate more additional training samples in \cite{wug2018fast}. When the first-frame ground-truth is provided in the form of bounding box, in \cite{wang2019fast}, original Siamese trackers is modified to generate the segmentation of the target object. In \cite{voigtlaender2019boltvos}, original R-CNN network is modified to a conditional R-CNN network, and a temporal consistency re-scoring algorithm is utilized to find candidate proposals of the target object, followed by a salience segmentation network to find the final result.

\subsection{Deep Reinforcement Learning}
Currently, RL has been applied to many computer vision applications. In the task of VOT, \cite{xiang2015learning} adopts RL to learn a similarity function for data association. In \cite{choi2018real}, RL is applied to choose the appropriate template from a template pool. In terms of VOS, Han et al. splits the VOS task into two sub-tasks including finding the optimal object box, and finding the context box  \cite{han2018reinforcement}. This work is desired by the fact that the obtained segmentation masks vary under different object boxes and context boxes for an identical frame. In this way, RL is naturally suitable to select the optimal object box and context box for each frame.  

\begin{figure*}[ht]
	\centering
	\includegraphics[width=1 \linewidth]{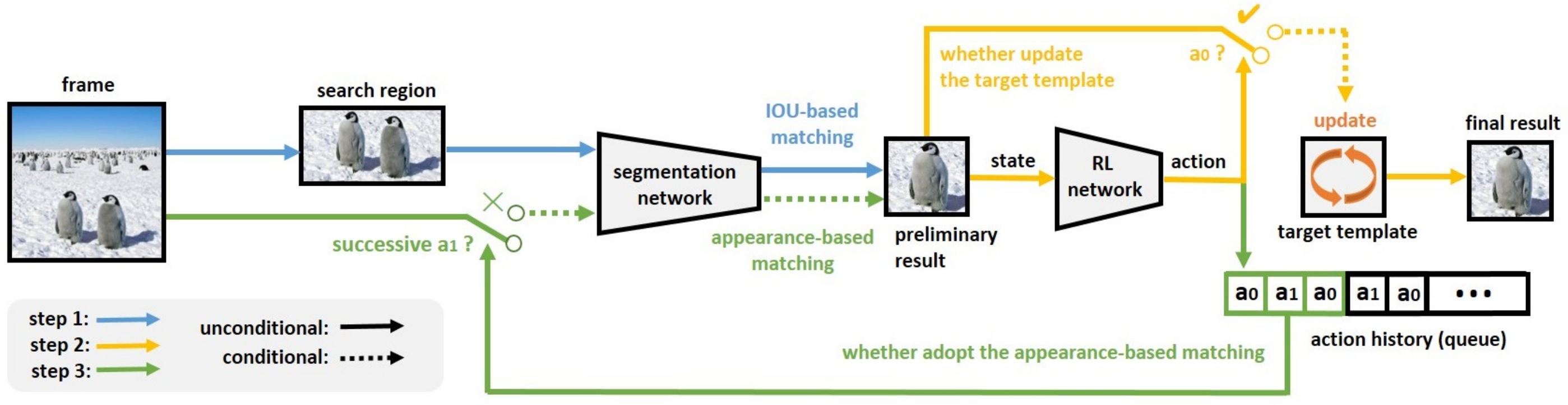}
	\caption{Architecture of our method where unconditional paths (full line) indicate they will be conducted in any situation, while conditional paths (dashed line) indicates they will be conducted only in some particular situations.}
	\label{framework}
\end{figure*}

\section{Our Approach}

\subsection{Overview}

Box-level semi-supervised VOS only provides the first-frame box-level ground-truth, instead of the first-frame pixel-level ground-truth. The main objective of our work is to utilize RL to boost the performance of box-level semi-supervised VOS in terms of both accuracy and running speed, by improving its matching mechanism.To do this, a RL agent is utilized to make two significant decisions simultaneously, including adopting which matching method, IOU-based matching or appearance-based matching, and whether to update the target template or not.

\begin{figure}[t]
	\begin{center}
		\includegraphics[width=0.6\linewidth]{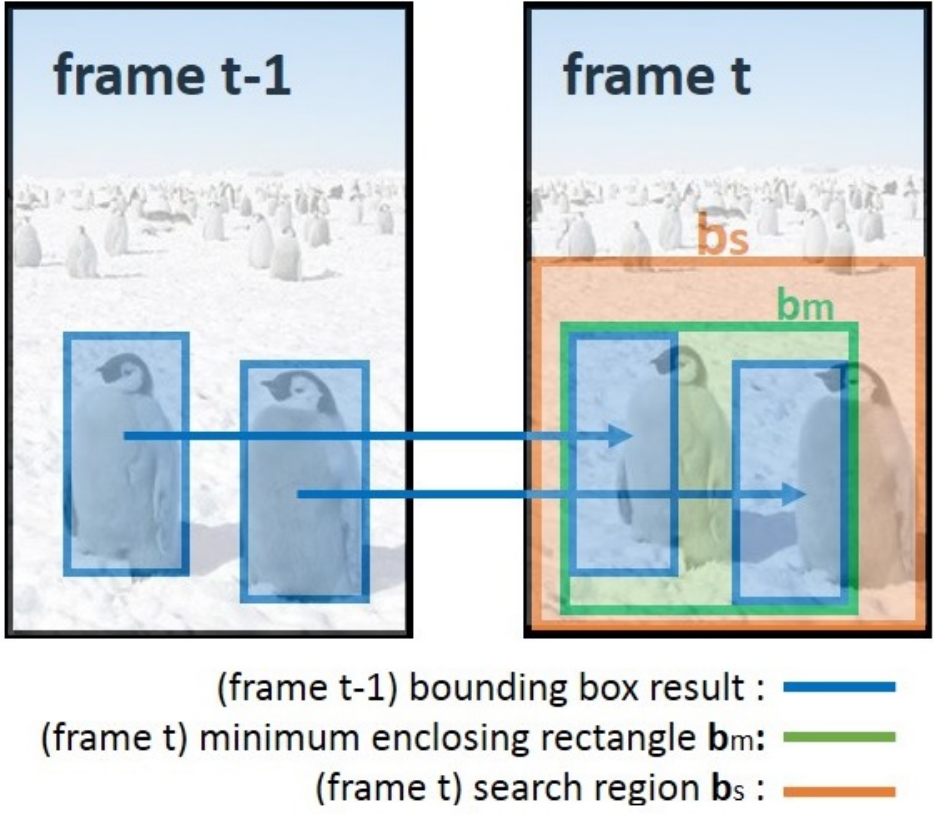}
	\end{center}
	\caption{Illustration of the minimum enclosing rectangle $b_m$ and search region $b_s$. $b_s$ is generated by $b_m$ by expansion.}
	\label{searchRegion}
\end{figure}

Specifically, as can be observed from Figure \ref{framework}, the processing of the current frame $f_t$ is split into three steps. The first step is to adopt the IOU-based matching to generate a temporary preliminary result. Specifically, the search region $b_s$ of the targets is determined first (see Figure \ref{searchRegion}), which will be fed into a general instance segmentation network (e.g. YOLACT \cite{daniel2019yolact}, Mask R-CNN \cite{he2017mask}) to generate numerous candidate predictions. Then, IOU-based matching will be adopted to find the preliminary result among all candidate predictions.

The second step is to determine the update mechanism for the target object information (target template) according to the correctness and quality of the preliminary result, which is judged by the RL agent. If it is good, the target template will be  entirely replaced by the preliminary result. Otherwise, the preliminary result will be discarded and the target template will keep unchanged. Ultimately, the final result is generated according to the target template.

The third step is to determine whether the appearance-based re-detection is essential for $f_t$. If the target is lost, in other words, the preliminary result keeps terrible for $N$ successive frames, the target needs to be re-detected again using the appearance-based matching. Otherwise, re-detection for $f_t$ is not needed, and the next frame $f_{t+1}$ will be processed. In terms of the appearance-based re-detection, the whole frame, rather than $b_s$, will be fed into a general instance segmentation network. Then, a new result will be selected by the appearance-based matching method among all candidate predictions. The second step is conducted again to generate a new final result. Note that the third step is conducted no more than once for each frame.    

\subsection{Agent Action}

A RL agent is used to address two complicated challenges during the matching process, which have been ignored by all existing detection-based VOS methods.

\begin{figure}[t]
	\begin{center}
		\includegraphics[width=0.9 \linewidth]{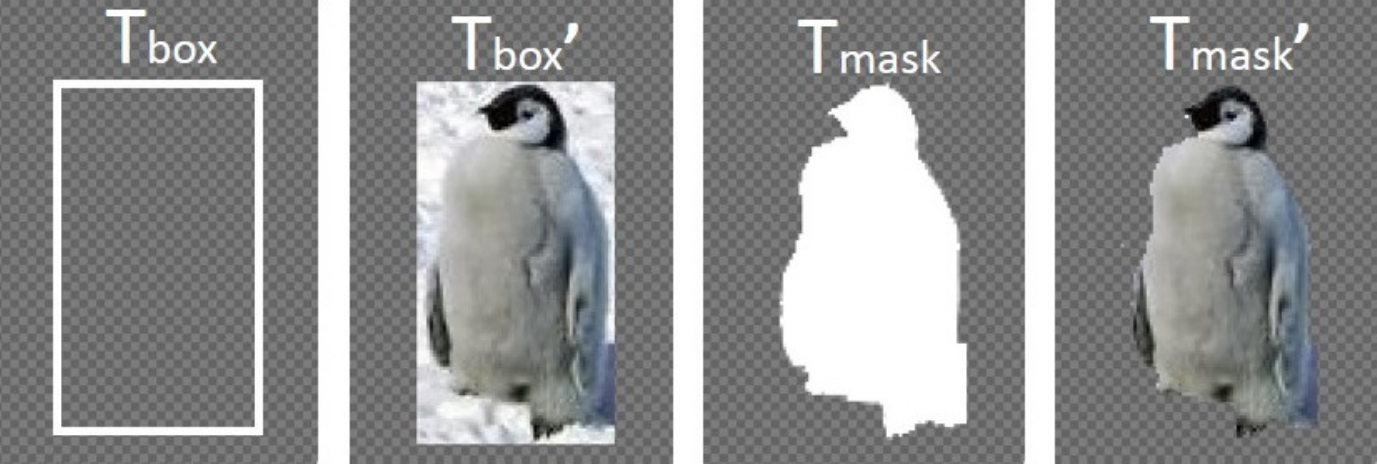}
	\end{center}
	\caption{Illustration of elements in the target template ($T_{box}$, $T_{mask}$, $T_{box'}$, $T_{mask'}$). Elements in the predicted result ($P_{box}$, $P_{mask}$, $P_{box'}$, $P_{mask'}$) are in the same form.}
	\label{template}
\end{figure}

In our approach, \emph{target template} is used to represent the target information, which is an important concept. As shown in Figure \ref{template}, the target template consists of the target's bounding box $T_{box}$, segmentation mask $T_{mask}$, cropped image $T_{box'}$ inside $T_{box}$, cropped image $T_{mask'}$ inside $T_{mask}$ and the whole frame $T_{frame}$. Correspondingly, the \emph{predicted result} incorporates the bounding box $P_{box}$, segmentation mask $P_{mask}$, cropped image $P_{box'}$ inside $P_{box}$, cropped image $P_{mask'}$ inside $P_{mask}$ and the current frame $P_{frame}$.

The first challenge is to decide whether to update the target template using the predicted result. Traditionally, the target template is updated in a rough way, without taking the correctness or quality of the predicted result into consideration. Therefore, when the segmentation network predicts a terrible result for current frame $f_t$, which may even refer to another object rather than the target, the target template will still be replaced by the incorrect result. This mortal error will make the tracker drift to the wrong target, causing a substantial accuracy drop. Note that this error cannot be avoided by adopting a better matching method because the target was decided before the matching process.

In this way, a ``smart switch'', which is able to decide whether to update the target template according to the quality of the predicted result, may be the best solution for this challenge. Rather than making the decision heuristically, we adopt a RL agent to make such a decision. The action set $A$ for the RL agent contains 2 candidate actions $a_i\in A$, including $a_0$ to replace the target template using the predicted result of $f_t$, and $a_1$ to ignore the predicted result of $f_t$ and keep the target template unchanged. 

The second challenge is to decide whether to adopt the fast IOU-based matching method or the accurate appearance-based matching method. In our approach, IOU-based matching views the candidate prediction with the highest IOU score as the correct one, written as:
\begin{equation} 
S_{IOU} = \alpha IOU(T_{box},P_{box}) + \beta IOU(T_{mask},P_{mask}),
\label{siou}
\end{equation}
\begin{equation}
\alpha + \beta = 1,
\label{ab}
\end{equation}
where $\alpha$ and $\beta$ refer to the weight of these two IOUs. 

Appearance-based matching views the candidate prediction with the highest appearance similarity as the correct one:
\begin{equation}
S_{a} = Similarity(T_{box'},P_{box'}),
\label{sapp}
\end{equation}
where $Similarity()$ is a Siamese style network, whose inputs are image patches within $T_{box'}$ and $P_{box'}$. Then, their individual embedding vectors are generated. A small L2 distance between these two vectors indicates two patches are similar, and vice versa.   

Having two matching mechanisms is inspired by the observation that the fast IOU-based matching performs well for most normal frames, while the appearance-based matching is only essential for a few difficult situations, especially when the target disappears and reappears again. Therefore, the selection of the matching method is pretty significant to trade off between running speed and accuracy. 

Instead of adding one more RL agent, we decide to choose the matching method according to the action history. In fact, the action history intrinsically indicates the prediction quality. If the RL agent predicts $a_1$ for $N$ successive frames, it is very likely that the target has been lost, and the appearance-based matching is essential to be adopted to detect the target on the whole frame. 

\subsection{State and Reward}

The state $s_t$ is the input of the RL agent for frame $f_t$, including the information assisting the RL agent to predict an optimal action $a_t$. 

In our approach, $s_t$ consists of two parts to provide sufficient information to the RL agent. The first part $S_T$ is the modified image of $T_{frame}$ where $T_{box'}$ remains unchanged while the area outside $T_{box}$ is blackened, written as:
\begin{equation} 
S_{T} = T_{box'}  \cup \Phi (\{i|i\in T_{frame}, i \notin T_{box'} \}),
\end{equation}
where function $\Phi$ is to set all pixels black. It provides both the location and appearance information of the target in $T_{frame}$.

The second part $S_P$ is the modified image of $P_{frame}$ where $P_{mask'}$ remains unchanged while the area outside $P_{mask}$ is blackened:
\begin{equation} 
S_{P} = P_{mask'}  \cup \Phi (\{i|i\in P_{frame}, i \notin P_{mask'} \}).
\end{equation}
It provides both the location and appearance information of the predicted objects, as well as its segmentation information. 

The ultimate $s_t$ is the concatenation of the feature maps of $S_T$ and $S_P$:
\begin{equation}
s_t = feature(S_T) + feature(S_P).
\end{equation} 

In details, we adopt Resnet-50 \cite{he2016deep}, pre-trained on the ImageNet classification dataset \cite{deng2009imagenet}, to extract the feature map of $S_T$ and $S_P$. We use the first 5 blocks of Resnet-50 \cite{he2016deep} which results in a feature map with the size of $\mathbb{R}^{1\times1\times2048}$ for both $S_T$ and $S_P$, and $s_t$ is with size $\mathbb{R}^{1\times1\times4096}$. Finally, $s_t$ will be fed into the RL agent to predict the action for frame $f_t$. 

The reward function, which reflects the accuracy of the final segmentation result for the video sequence, is defined as $ r_{t}=g(s_{t},a_t)$:

\begin{equation} g(s_t,a)=\left\{\begin{matrix}
100J_t^3+10 & J_t>0.1 \\ 
-10 & J_t\leq0.1 
\end{matrix}\right.
\label{reward},
\end{equation}
where $J_t$ refers to the IOU between $P_{mask}$ and the ground-truth mask. Using the cube of $J_t$ expands the difference between the good action's reward and the bad action's reward, which helps to speed up the training of the RL agent.

\subsection{Search Region Size}

The size of the search region $b_s$ greatly affects the quality of the segmentation result. As can be observed from Figure \ref{searchRegion}, $b_s$ is generated using $b_m$ (minimal box covering all target objects in the previous frame) by expansion. The expansion ratio of $b_m$ varies according to the video's characteristic. In detail, there are three expansion ratio from $b_m$ to $b_s$, including a big one, a small one and an intermediate one. First, each target's displacement distance  between two adjacent frames is calculated. If anyone is larger than a threshold, the big expansion ratio is selected. Otherwise, if two target objects are close to each other (IoU between their bounding boxes is not zero), the small expansion ratio is chosen. If not, the intermediate ratio is selected. 	

\subsection{Actor-Critic Training}

In our approach, the RL agent is trained under the ``actor-critic'' framework \cite{konda2000actor}, which is a prevalent RL framework consists of two sub-networks including an ``actor'' sub-network to generate the action and a ``critic'' sub-network to check the quality of this action. Once the RL agent is fully trained, only the ``actor'' sub-network is used during the inference time. 

In our ``actor-critic'' framework, given a current frame $f_t$, the first step is to feed the state $s_t$ into the ``actor'' network and generate an action $a_t$ to decide whether to update the target template using the predicted result. The corresponding reward $r_t$ will also be obtained after conducting this action. $r_t$ is calculated by the region similarity $J_t$ according to (\ref{reward}).

Our ``critic'' network will be trained in the value-based way. Specifically, the parameters are updated as follows

\begin{equation}
w = w{}' + \l_c \delta_t \nabla_{w{}'} V_{w{}'}(s_t), 
\label{criticupdate}
\end{equation}
where
\begin{equation}
\delta_t =  r_t+ \gamma V_{w{}'}(s_{t+1})-V_{w{}'}(s_t).
\label{tderror}
\end{equation}

In (\ref{criticupdate}) and (\ref{tderror}), $w$ and $w{}'$ indicate the weight of the ``critic'' model after and before update.  $l_c$ is the learning rate of the ``critic'' model. $\delta_t$ is the TD error which indicates the difference of the actual score and the predicted score. $V_{w{}'}(s_t)$ refers to the accumulated reward of state $s_t$ which is predicted by the ``critic'' model before update. $\gamma$ refers to the discount factor.

The ``actor'' network will be updated after the ``critic'' network in a policy-based way, as follows

\begin{equation} 
\theta  = \theta{}' + \l_a   \nabla (log\pi_{\theta{}'} (s_{t},a_{t}))   A(s_{t},a_{t}), 
\label{actor}
\end{equation}
where
\begin{equation}
A(s_{t},s_{t}) = Q(s_{t},a_{t})-V(s_{t}) = \delta_t. 
\label{advantage}
\end{equation}

In (\ref{actor}) and (\ref{advantage}), $\theta$ and $\theta{}'$ indicate the weight of the ``actor'' model after and before update. $l_a$ is the learning rate of the ``actor'' model. $\pi(s,a)$ is the policy function which indicates the probability of selecting action $a$ in state $s$. $V(s_t)$ is the score of the state $s_t$. $Q(s_t,a_t)$ is the score of the state $s_t$ if the action $a_t$ is executed. $A(s,a)$ refers to the advantage function.

In this way, our ``actor-critic'' framework avoids the disadvantages of both value-based and policy-based methods during the training process. In other words, our RL agent is allowed to be trained and updated at each frame, rather than waiting until the end of the episode, which dramatically speeds up the training process yet maintains training stability.
\begin{table*}[]
	\caption{Quantitative comparison with other methods on the DAVIS 2017 (Da 17), DAVIS 2016 (Da 16), SegTrack V2 (ST) and Youtube-Object (YOs) datasets, measured by the mean region similarity ($J$), as well as the average score of region similarity and boundary similarity ($J\&F$). \textbf{FT} indicates fine-tuning, \textbf{M} indicates using the first-frame masks, t(s) indicates the average running time per frame in seconds. The method with the best score is bold, and the method with the second best score is marked in underline.}
	\begin{center}
		\begin{tabular}{|p{60pt}  p{10pt}  p{10pt} | p{20pt}<{\centering} | p{40pt}<{\centering} | p{55pt}<{\centering} | p{40pt}<{\centering} | p{55pt}<{\centering} | p{35pt}<{\centering} | p{40pt}<{\centering}|}
			\hline
			Method  &  FT  &  M &  t(s)  &  Da 17 - $J$  &  Da 17 - $J\&F$  & Da 16 - $J$ & Da 16 - $J\&F$ & ST - $J$  &  YOs - $J$  \\
			\hline\hline
			Ours           &\ding{56} &\ding{56}  &\underline{0.09}  &\textbf{69.1}   &\underline{70.6}         &\underline{77.5}         &\underline{78.9} &\textbf{79.2} &\textbf{79.3} \\
			BoLTVOS\cite{voigtlaender2019boltvos}        &\ding{56} &\ding{56}  &1.45              &\underline{68.4}&\textbf{71.9}         &\textbf{78.1}         &\textbf{79.6}          &-             &-        \\
			BoLTVOS *      &\ding{56} &\ding{56}  &1.45              &60.9            &64.9         &\textbf{78.1}         &\textbf{79.6}          &-             &-        \\
			SiamMask\cite{wang2019fast}       &\ding{56} &\ding{56}  &\textbf{0.06}     &54.3            &55.8         &71.7         &69.8          &-             &-        \\
			SiamMask *     &\ding{56} &\ding{56}  &0.11              &59.5            &63.3         &75.6         &75.9          &-             &-        \\
			MSK(box)\cite{perazzi2017learning}       &\ding{56} &\ding{56}  &12                &-               &-            &73.7         &-             &\underline{62.4}          &\underline{69.3}     \\
			\hline\hline
			STM\cite{oh2019video}            &\ding{56} &\ding{52}  &0.16              &69.2            &74.1         &84.8         &88.1          &-             &-       \\
			FEELVOS\cite{voigtlaender2019feelvos}        &\ding{56} &\ding{52}  &0.51              &69.1            &71.5         &81.5         &81.8          &-             &78.9     \\
			RGMP\cite{wug2018fast}           &\ding{56} &\ding{52}  &0.28              &64.8            &66.7         &81.1         &81.7          &71.7          &-        \\
			VideoMatch\cite{hu2018videomatch}     &\ding{56} &\ding{52}  &0.35              &56.5            &62.4         &81.0         &80.9          &79.9          &-        \\
			OSMN\cite{yang2018efficient}           &\ding{56} &\ding{52}  &0.28              &52.5            &54.8         &74.0         &73.5          &-             &69.0     \\
			\hline\hline
			PReMVOS\cite{luiten2018premvos}        &\ding{52} &\ding{52}  &37.21             &73.9            &77.8         &85.6         &86.5          &-             &-        \\
			OSVOS-S\cite{maninis2017video}        &\ding{52} &\ding{52}  &9                 &64.7            &68.0         &84.9         &86.8          &-             &83.2     \\
			OnAVOS\cite{voigtlaender2017online}         &\ding{52} &\ding{52}  &26                &61.0            &63.6         &85.7         &85.0          &66.7          &77.4     \\
			CINM\cite{bao2018cnn}           &\ding{52} &\ding{52}  &$>$120            &64.5            &67.5         &83.4         &84.2          &77.1          &78.4     \\
			\hline
		\end{tabular}
	\end{center}
	
	\label{comparison}
	
\end{table*}

\section{Implementation Details}

\subsection{Segmentation Network Training}
In our approach, the training of the instance segmentation network follows the strategy of YOLACT \cite{daniel2019yolact}. The first step is to pre-train a ResNet-101 network \cite{he2016deep} using the ImageNet classification dataset \cite{deng2009imagenet}. Then, this network with FPN \cite{lin2017feature} is used as the feature backbone for the segmentation network. Finally, the segmentation network will be trained on the PASCAL VOC dataset \cite{everingham2010pascal} with three losses including the classification loss, box regression loss and the mask loss calculated by the pixel-level binary cross-entropy between the predicted masks and the ground-truth.

\subsection{RL Agent Training}

Our RL agent is trained on the training set of the DAVIS 2017 dataset where all video sequences of the training set are divided into video clips with the fixed number of frames in advance. A video clip, consisting of 10 consecutive frames, is used as an episode for the training of the RL agent. 20 video clips will be randomly selected as a batch. At the beginning, the learning rate $\l_a$ for the ``actor'' model is 1e-4, and the learning rate $\l_c$ is 5e-4. $\l_a$ and $\l_c$ decrease gradually during the training, and they decrease by 1\% for each 200 iterations. The discount rate $\gamma$ for the reward is 0.9. The training of our RL agent takes about 10 days on a NVIDIA GTX 1080 Ti GPU and a 12 Core Intel i7-8700K CPU@3.7GHz. 

In terms of other hyper-parameters, when calculating the $S_{IOU}$ in (\ref{siou}), we found $\alpha = 1$, $\beta = 0$ for the first frame, and $\alpha = 0.5$, $\beta=0.5$ for other frames work well, because the pixel-level ground-truth is not available for our task. For the appearance-based matching, we found it is better to re-detect the target using the appearance-based matching when action $a_1$ is taken for 3 successive frames.

\section{Experiments}

\subsection{Experiment Setup}
We split the experimental evaluation into two sections including the evaluation on the VOS dataset and the evaluation on the VOT dataset.

For the VOS experiments, we evaluate our method on four widely-used datasets including DAVIS 2017 dataset \cite{pont20172017}, DAVIS 2016 dataset \cite{perazzi2016benchmark}, Youtube-Object dataset \cite{prest2012learning}, and Segtrack V2 dataset \cite{li2013video}. DAVIS 2016 dataset consists of 50 high quality videos and 3,455 frames, with 30 videos for training and 20 videos for evaluation. In DAVIS 2016 dataset, only a single target is annotated per video sequence. DAVIS 2017 dataset extends DAVIS 2016 dataset, consisting of 60 video sequences for training and 30 video sequences for evaluation, spanning multiple occurrences of common video object segmentation challenges such as occlusions, motion-blur and appearance changes. In DAVIS 2017 dataset, each video sequence contains 2.03 object on average, and a maximum of 5 objects to be tracked in a single video sequence. In Youtube-object, there are 155 video sequences and a total of 570,000 frames. All these video sequences are divided into 10 sets according to the category. Youtube-Object dataset does not split the training set and the evaluation set, so we set all video sequences as the evaluation set. Note that Youtube-Object is not an instance-level dataset, in other words, for some videos, several individual targets are annotated into one object as the foreground, which does not completely match our task. Therefore, we split the annotations of these video sequences, so that each target owns its individual instance-level annotation. In SegTrack V2 dataset, there are 14 video sequences with more occlusion and appearance changes compared with Youtube-Object dataset. 
Originally, these datasets only provide the pixel-level ground-truth, which does not match our task. Therefore, we generate the bounding boxes according to the pixel-level ground-truth in advance as the first-frame box-level ground-truth.

We valuate our method following the approach proposed in~\cite{pont20172017}. The adopted evaluation metrics include region similarity $J$ and contour accuracy $F$. The region similarity is calculated as $J=\left | \frac{m\cap gt}{m\cup gt} \right |$ by the intersection-over-union between the predicted segmentation $m$ and the ground-truth $gt$. The contour accuracy is defined as $F=\frac{2P_cR_c}{P_c+R_c}$, which indicates the trade-off between counter-based precision $P_c$ and recall $R_c$ using the F-measure.

For the VOT experiments, we evaluate our VOS method on the LTB35 dataset \cite{kristan2018sixth} which is a long-term VOT dataset and was adopted to evaluate the long-term tracking performance in the VOT2018 challenge \cite{kristan2018sixth}. LTB35 dataset consists of 35 video sequences with 4,200 frames for each video sequence on average. In addition, the target will disappear and reappear again for 12.4 times, with an average target absence period of 40.6 frames per video. In this way, this dataset is suited to check the algorithm's ability to re-detect the disappeared target. We evaluate our method following the standard metric for LTB35 dataset. The performance is measured by the $F$ score which is calculated as $F = \frac{2P_rR_e}{(P_r+R_e)}$, where $P_r$ indicates the precision and $R_e$ indicates the recall. Algorithms will be ranked by the maximum $F$ score under different confidence thresholds. 

\begin{figure}[t]
	\begin{center}
		\includegraphics[width=0.9 \linewidth]{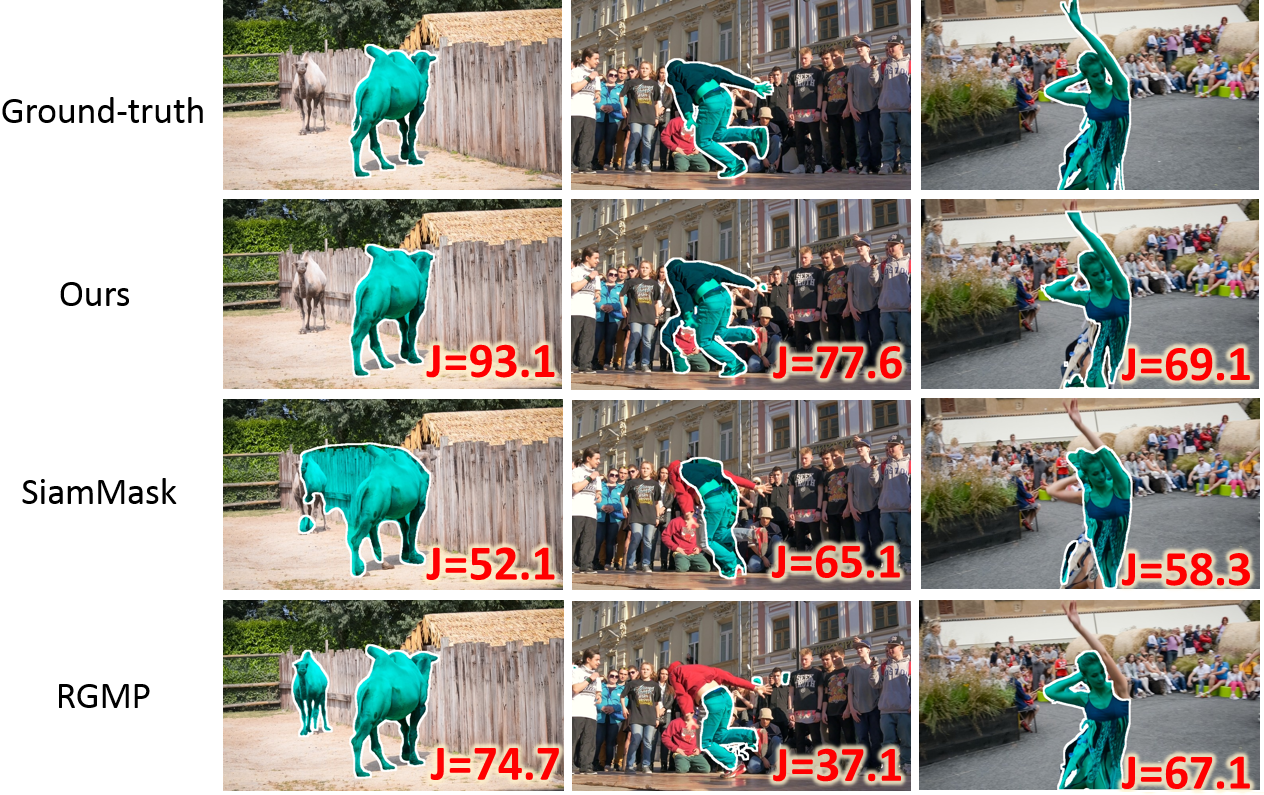}
	\end{center}
	\caption{Visualization results of different methods.}
	\label{visualization}
\end{figure}

\subsection{Comparison with State-of-the-arts}
For the experimental evaluation on the VOS dataset, we compare our method with other state-of-the-art VOS methods, which are classified into three groups. Methods in the first group only use the first-frame box-level ground-truth, including BoLTVOS \cite{voigtlaender2019boltvos}, SiamMask \cite{wang2019fast} and MSK \cite{perazzi2017learning}. Methods in the second group adopt the first-frame pixel-level ground-truth but do not fine-tune on it, including STM \cite{oh2019video}, FEELVOS \cite{voigtlaender2019feelvos}, RGMP \cite{wug2018fast}, VideoMatch \cite{hu2018videomatch} and OSMN \cite{yang2018efficient}. Methods in the third group fine-tune on the first-frame pixel-level ground-truth, including PReMVOS \cite{luiten2018premvos}, OSVOS-S \cite{maninis2017video}, OnAVOS \cite{voigtlaender2017online}, and CINM \cite{bao2018cnn}. All quantitative results of the comparison are summarized in Figure \ref{tradeoff} and Table \ref{comparison}. In Table \ref{comparison}, the method with the highest score is highlighted and the method with the second best score is marked with underline.

As can be observed from Table \ref{comparison}, for the evaluation of the DAVIS 2017 dataset, compared with other methods which only rely on the first-frame box-level ground-truth (BoLTVOS* removes the re-scoring network and SiamMask* adopts the Box2Seg network of BoLTVOS), our method is virtually 15 times faster the previous state-of-the-art method BoLTVOS \cite{voigtlaender2019boltvos}, and our accuracy (mean region similarity $J_m$) is even higher than BoLTVOS \cite{voigtlaender2019boltvos} at the same time. For another competitive method SiamMask \cite{wang2019fast}, which runs virtually as fast as our method, our accuracy is much higher than it by around 15\%. In addition, in the second group, the proposed method approximately achieves the same $J_m$ as the state-of-the-art method STM\cite{oh2019video}, trained without the additional Youtube-VOS dataset \cite{xu2018youtube}, as it was not used in our method. When compared with the methods in the third group, our method also outperforms most of these methods. Only PReMVOS \cite{luiten2018premvos} achieves a higher accuracy than our method, but our method runs 370 times faster than it. For the evaluation of DAVIS 2016 dataset, as it is not an instance-level dataset, our method ranks the second among methods in the first group. For the evaluation for SegTrack V2 dataset and Youtube-Object dataset, our methods also achieves a competitive result even compared with methods using the first-frame pixel-level ground-truth. Some visualization results are shown in Figure \ref{visualization}.

\begin{figure}[t]
	\begin{center}
		\includegraphics[width=0.8 \linewidth]{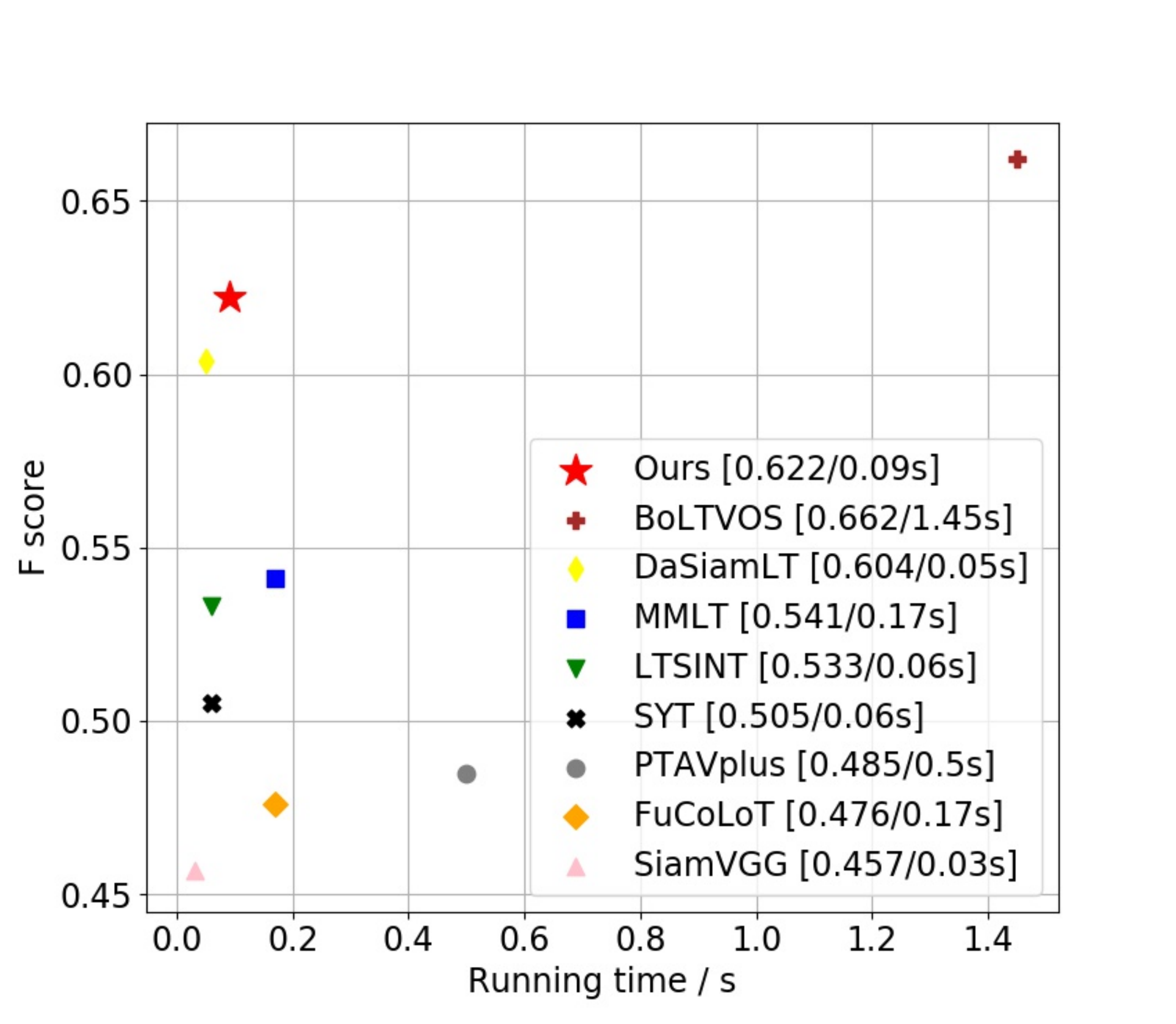}
	\end{center}
	\caption{Speed-accuracy trade-off on VOT2018 LT dataset.}
	\label{vot_tradeoff}
\end{figure}

For the experimental evaluation of VOT, the comparison with other state-of-the-art methods \cite{kristan2018sixth} is conducted on the VOT 2018 long-term dataset. As can be observed from Figure \ref{vot_tradeoff}, our method achieves an F score of 0.622, which is quite competitive compared with other VOT methods. Overall our method achieves very good speed accuracy trade-off for the VOT task. This result also shows our method is able to handle both VOS and VOT tasks, even in the situations where the target disappears and reappears again at a different place.

\subsection{Ablation studies}

\begin{table}[]
	\caption{Ablation studies on the DAVIS 2017 dataset, measured by the mean region similarity ($J_m$).}
	\begin{center}
		\begin{tabular}{|p{110pt}<{} |p{30pt}<{\centering}|p{30pt}<{\centering}|}
			\hline
			Method & $J_m$ & t(s)\\
			\hline\hline
			no update                  & 27.3     & 0.03  \\
			simple update (IOU)        & 63.1     & 0.03  \\
			simple update (appearance) & 67.2     & 1.10  \\
			RL update w/o re-detection & 68.1     & 0.06  \\
			Ours                       & \textbf{69.1}     & 0.09  \\
			\hline\hline
   			Ours (supervised)           & 60.2     &0.09   \\
			\hline\hline
			BoLTVOS                    & 68.4     & 1.45  \\
			SiamMask                   & 54.3     & 0.06  \\
			\hline
		\end{tabular}
	\end{center}
	\label{ablation}
\end{table}

\textbf{Contribution of each component:} We conduct ablation studies on DAVIS 2017 \cite{pont20172017}, where parts of our methods are disabled to investigate the contribution of each component.

First, we totally remove the target template update mechanism, obtaining the method \textbf{no update}, which cause numerous drift, and leads to a terrible accuracy (27.3\%). Then, we evaluate the simple update mechanisms, where the target template will always be updated, and only IOU-based matching or only appearance-based matching is adopted to select the predicted result, obtaining method \textbf{simple update (IOU)} and method \textbf{simple update (appearance)} respectively. As can be observed from Table \ref{ablation}, only adopting IOU-based matching already achieves an acceptable accuracy (63.1\%). Although the accuracy of method \textbf{simple update (appearance)} is 4.1\% greater than method \textbf{simple update (IOU)}, the sacrifice on speed is unacceptable (from 0.03s to 1.1s), which proves the inefficiency of the simple target template update mechanism, where only appearance-based matching is adopted. In addition, the accuracy of method \textbf{simple update (appearance)} is still lower than that of our overall method, which demonstrate that the target template update issue cannot be totally solved simply by adopting an accurate matching method. Finally, we evaluate our method without the usage of the appearance-based matching for re-detection, obtaining method \textbf{RL update w/o re-detection}, whose accuracy is 1.0\% lower than that of our overall method. This gap is not big, because the situation where the target disappears and reappears from another place is pretty rare in the DAVIS 2017 dataset. 

\textbf{RL or supervised learning?} Apart from training the network with RL, we also attempt to train the network in the supervised way, i.e. evaluating the fixed label for each frame in advance before training, but finally, as can be observed from Table \ref{ablation}, we find the model trained under the reinforcement learning way performs much better than the model trained under the supervised way, which achieve the accuracy of 69.1\% (RL) and 60.2\% (supervised). We believe the major reason is that, a RL model considers not only the current profit but also the potential profit in the future, due to the adopted accumulative future reward for training. In other words, the model trained in a supervised way tends to be myopic, while the model trained in the RL way pays more attention to the global and overall performance, making it more suited to video-related tasks.

\textbf{Choice of instance segmentation:} Apart from adopting the one-stage segmentation network, like YOLACT \cite{daniel2019yolact}, we also attempt to adopt the two-stage segmentation network, like Mask R-CNN \cite{he2017mask}. Then, the running speed of our method drops slightly from around 90ms to around 150ms, and it is still around 10 times faster than the previous state-of-the-art method, BoLTVOS \cite{voigtlaender2019boltvos}. As Mask R-CNN achieves higher accuracy than YOLACT \cite{daniel2019yolact}, our final accuracy is even slightly higher than the proposed one. Note that the choice of the instance segmentation method does not greatly affect the final result, both for running speed and accuracy.

\section{Conclusion}

In this paper, an RL-based template matching and updating mechanism is proposed to handle box-level semi-supervised VOS. A single RL agent is applied to make these decisions jointly, which is trained using an actor-critic RL framework. Evaluation on common datasets for both VOS and VOT demonstrates the great performance of our method. In the future, we plan to design more matching mechanisms and template target update mechanisms to further improve the performance of our method.

{\small
\bibliographystyle{ieee_fullname}
\bibliography{egbib}
}

\end{document}